# Computational Aspects of the Möbius Transform

Robert KENNES and Philippe SMETS*

IRIDIA, Université Libre de Bruxelles
Av. F. D. Roosevelt 50 - CP 194/6
B-1050 Brussels - Belgium
E-mail : r01505@bbrbfu01.bitnet

**Abstract**
In this paper we associate with every (directed) graph G a transform called the *Möbius transform* of the graph G. The Möbius transform of the graph $(\wp\Omega,\subseteq)$ is of major significance for Dempster-Shafer theory of evidence. However, because it is computationally very heavy, the Möbius transform together with Dempster's rule of combination is a major obstacle to the use of Dempster-Shafer theory for handling uncertainty in expert systems. The major contribution of this paper is the discovery of the 'fast Möbius transforms' of $(\wp\Omega,\subseteq)$. These 'fast Möbius transforms' are the fastest algorithms for computing the Möbius transform of $(\wp\Omega,\subseteq)$. As an easy but useful application, we provide, via the commonality function, an algorithm for computing Dempster's rule of combination which is much faster than the usual one.

## 0. Introduction and motivations

Let $\Omega$ be a finite non empty set and $\wp\Omega$ be its power set equipped with the inclusion relation. In the Dempster-Shafer theory of evidence - the standard reference of which is [Shafer 76], see also [Smets 88] - a *basic belief assignment* (bba) on $\Omega$ is any function

$$m : \wp\Omega \to [0\ 1] \text{ such that } \sum_{X \in \wp\Omega} m(X) = 1$$

m(X) is the *mass* of X. Most often it is also required that m(Ø)=0. Anyway, any basic belief assignment m determines its *belief function* $\text{bel}_m : \wp\Omega \to [0\ 1]$ defined by

$$\forall A \in \wp\Omega : \text{bel}_m(A) = \sum_{X \subseteq A, X \neq \emptyset} m(X)$$

$\text{bel}_m(A)$ is the *belief* of A induced by the bba m. If, more generally, we consider any function $m: \wp\Omega \to \mathbf{R}$, the previous formula defines the functional

$$\mathbf{R}^{\wp\Omega} \to \mathbf{R}^{\wp\Omega} : m \to \text{bel}_m$$

where $\mathbf{R}^{\wp\Omega}$ denotes, as usual, the set of functions from $\wp\Omega$ to the set of real numbers. The notation $\text{bel}_m$, although rather explicit, does not do justice to the most important protagonist of the formula, that is the binary relation $\{(X,Y) \in \wp\Omega \times \wp\Omega \mid X \neq \emptyset, X \subseteq Y\}$. Thus the above functional, which will be called the *Möbius transform*, is induced by the relation $\{(X,Y) \in \wp\Omega \times \wp\Omega \mid X \neq \emptyset, X \subseteq Y\}$, or less exactly by the finite boolean lattice $(\wp\Omega,\subseteq)$.

The first section of the paper begins with the definition of the Möbius transform induced by an arbitrary graph. The Möbius transform defines in an obvious way a map between two categories. The just defined map is not a functor, but by generalizing the Möbius transform induced by a graph to the Möbius transform induced by a weighted graph, the map becomes a functor. Such a generalization sheds some light on the preceding situation by providing a recursive formula for computing the Möbius transform. The fundamental fact is that recursion is neither on the set $\Omega$ nor on the power set $\wp\Omega$ but on the inclusion relation. In both situations, graphs and weighted graphs, we define what we call M-algorithms: since a graph determines a functional, a sequence of graphs determines the composite of the functionals induced by each graph of the sequence. A natural problem is then to decompose a graph into subgraphs in order to get various algorithms computing the Möbius transform induced by the graph. In the second section, as an application of that decomposition, we provide 'fast' M-algorithms for computing the Möbius transform of $(\wp\Omega,\subseteq)$. In the third section we define the computational complexity of M-algorithms and in the fourth section we show that the previously defined 'fast' M-algorithms are the fastest among all M-algorithms computing the same Möbius transform. In the fifth part, as an application, we compute Dempster's rule of combination in a much faster way than the usual one.

Lydia Kronsjö points out, in her book [Kronsjö 85 p.20], that *efficient algorithms for solving the problems of arithmetic complexity are frequently based on a technique known as recursion*. She mentions that during the 1960's three very surprising algorithms were discovered: for the multiplication of two integers, for computing the discrete Fourier transform, and for the product of two matrices. As a matter of fact all these efficient algorithms are based on recursive formulas. The present paper is in keeping with this observation.

Due to lack of space, no proof will be given in this paper. All theorems are proved in [Kennes 90].

## 1. The Möbius functor

### 1.1 How graphs operate on functions

Let S and T be finite sets. A subset G of the cartesian product $S \times T$ is called a *(directed) graph* from S to T. We write indifferently $G: S \to T$ or $G \subseteq S \times T$. Sometimes we will say *arrows of G* instead of *ordered pairs of G*. When no confusion is possible we use the same symbol to denote a binary relation and the set of ordered pairs it determines on a particular set. Explicitly, if R is a binary relation, the graph $((s,t) \in S \times S \mid sRt)$ it determines on the set S will also be denoted by R. Throughout this paper all graphs are finite.

SET denotes the category of sets. FGRAPH denotes (confusingly!) the category of which the objects are the *finite sets* and the arrows are the *graphs* $G: S \to T$ together with the usual *composition of graphs*.

Any arrow of the category FGRAPH, i.e. any graph, determines a *Möbius transform*. More explicitly:

**Definition 1.** The graph $G: S \to T$ determines the functional:
$$M^G: R^S \to R^T: f \to M^G(f) = f^G$$
defined by
$$\forall t \in T: f^G(t) = \sum_{sGt} f(s) = \sum_{s \in G^{-1}(t)} f(s)$$

We call $M^G$ the *Möbius transform of G*.

(We recall that : $\sum_{s \in \emptyset} f(s) = 0$.)

An extensive reference to (a special case of) the Möbius transforms is [Aigern 79, chap. IV], where it is called the *sum function*.

Since the graph G is finite the sum is well defined. As a matter of fact we might have taken *lower semi-finite graphs* i.e. graphs such that all the sets $G^{-1}(t)$ are finite. But we will stick to finite graphs.

$M^G$ may be interpreted as follows. It transforms any *distribution mass* on S into a *distribution mass* on T in the following way: all the masses f(s) are dragged along (or transferred by) the arrows $(s,t) \in G$ to the targets and they are added together at each target. $M^G$ may also be seen as a discrete analogue of the indefinite integral in calculus. We say equivalently that: G *computes* $M^G$ or that G is a *M-algorithm* of $M^G$ or that $M^G$ is the *Möbius transform* of G (or *induced by* G or *determined by* G or *associated with* G). For obvious reasons, which will appear later, the graph G is called the *obvious* M-algorithm computing $M^G$.

**Theorem 1.** The map M from the category FGRAPH to the category SET, M: FGRAPH $\to$ SET:
$(S \to M(S) = R^S, G \to M(G) = M^G)$ verifies:
(1) $M(1_S) = 1_{M(S)}$,
(2) For every graphs G, H: $S \to T$: $G = H$ iff $M(G) = M(H)$.

In general $M(G_2 \circ G_1) \neq M(G_2) \circ M(G_1)$, so M is not a functor FGRAPH $\to$ SET. Fortunately, by slightly modifying the category FGRAPH, M becomes a functor as it will be seen in the next section. This appears to be a key fact for our concern: this functor will provide a criterion for the equality $M(G_2 \circ G_1) = M(G_2) \circ M(G_1)$. The property (2) (*faithfulness* of M) implies that a Möbius transform is induced by exactly one graph.

Any finite sequence G of *queueing* graphs :
$$S_0 \xrightarrow{G_1} S_1 \xrightarrow{G_2} \ldots \xrightarrow{G_n} S_n$$
where the sets $S_0, S_1, \ldots, S_n$ are finite but not necessarily disjoint from each other, is called a *M-algorithm* (of length n). (To be brief, we will say *sequence of graphs* instead of *sequence of queueing graphs*.)

We say that the previous M-algorithm G computes the composite of
$$R^{S_0} \xrightarrow{M^{G_1}} R^{S_1} \xrightarrow{M^{G_2}} \ldots \xrightarrow{M^{G_n}} R^{S_n}$$
which is equal to $M^{G_n} \circ M^{G_{n-1}} \circ \ldots \circ M^{G_1}$, but not always equal to $M^{G_n \circ G_{n-1} \circ \ldots \circ G_1}$.

Let us recall that the *composite* of a sequence G
$$S = S_0 \xrightarrow{G_1} S_1 \xrightarrow{G_2} \ldots \xrightarrow{G_n} S_n = T$$
denoted by $C(G): S \to T$
is the graph $C(G) = G_n \circ \ldots \circ G_1: S \to T$.

In the next section, we show that if a sequence G computes the Möbius transform of a graph, then the graph is C(G), that is, if a sequence G computes the Möbius transform of a graph, then it computes the Möbius transform of $G_n \circ G_{n-1} \circ \ldots \circ G_1$.

### 1.2. How weighted graphs operate on functions

**Definition 2.** A *weighted graph* $\alpha: S \to T$ from S to T is a function: $\alpha: S \times T \to R: (s,t) \to \alpha(s,t)$.

[Aigner 79] is an extensive reference to weighted graphs or *incidence functions* as they are called there. Actually $\alpha$ can also be seen as a matrix $(\alpha(s,t))_{(s,t) \in S \times T}$.

The product of weigthed graphs is defined as follows:

**Definition 3.** If $\alpha: S \to T$ and $\beta: T \to U$ are two weighted graphs, their *product* $\alpha*\beta$ is defined by

$$\alpha*\beta\ (s,u) = \sum_{t \in T} \alpha(s,t) \cdot \beta(t,u)$$

This is in fact the product of the matrices $\alpha$ and $\beta$.

Each set S defines its identity weighted graph $\delta_S$, the *Kronecker function of S* [Aigner 79 p.140], which is the characteristic function of $1_S$ as a subset of $S \times S$. The $\delta_S$'s are the identities of the product. As a consequence, we get the category **WGRAPH** the objects of which are the *finite sets* and the arrows are the *weighted graphs* $\alpha: S \to T$ together with the *product*.

In the same way we defined the Möbius transform of a graph we define the Möbius transform of a weighted graph.

**Definition 4.** The weighted graph $\alpha: S \to T$ determines the following functional

$$M^\alpha: \mathbf{R}^S \to \mathbf{R}^T : f \to M^\alpha(f) = f^\alpha$$

defined by

$$\forall t \in T : f\alpha(t) = \sum_{s \in S} f(s) \cdot \alpha(s,t)$$

$M^\alpha$ will be called the *Möbius transform* of $\alpha$.

This is the product of the column-vector $(f(s))_{s \in S}$ by the matrix $(\alpha(s,t))_{(s,t) \in S \times T}$.

$M^\alpha$ may be seen as a discrete analogue of the Stieltjes indefinite integral in calculus and admits the same interpretation as $M^G$ where the $\alpha(s,t)$ are scaling factors. Now, M turns out to be a functor.

**Theorem 2.** The map $M : \mathbf{WGRAPH} \to \mathbf{SET}$:

$(S \to M(S) = \mathbf{R}^S, \alpha \to M(\alpha) = M^\alpha)$

from the category of weighted graphs to the category of sets is a functor we call the *Möbius functor*. This functor is faithful.

As in the case of graphs, we say that the following sequence of *queueing* weighted graphs

$$S_0 \xrightarrow{\alpha_1} S_1 \xrightarrow{\alpha_2} \ldots \xrightarrow{\alpha_n} S_n$$

is a *M-algorithm* of weighted graphs (of length n) which *computes* the composite functional of

$$\mathbf{R}^{S_0} \xrightarrow{M^{\alpha_1}} \mathbf{R}^{S_1} \xrightarrow{M^{\alpha_2}} \ldots \xrightarrow{M^{\alpha_n}} \mathbf{R}^{S_n}$$

In this case theorem 2 shows that the composite functional is also the Möbius transform of $\alpha_1*...*\alpha_n: S_0 \to S_n$. The following theorem is then simply another phrasing of the preceding theorem:

**Theorem 2'.** The following sequence of weighted graphs

$$S = S_0 \xrightarrow{\alpha_1} S_1 \xrightarrow{\alpha_2} \ldots \xrightarrow{\alpha_n} S_n = T$$

computes the Möbius transform of (only of) $\alpha_1*...*\alpha_n$.

We now provide a link between the two categories **FGRAPH** and **WGRAPH**.

Each graph $G \subseteq S \times T$ determines its weighted graph $\zeta_G$, *the zeta-function of G* [Aigner 79 p.140], which is its characteristic function (except for its codomain) as a subset of $S \times T$:

$$\zeta_G: S \times T \to \mathbf{R} : (s,t) \to \zeta_G(s,t)$$

$\zeta_G(s,t) = 1$ iff $(s,t) \in G$ and, $\zeta_G(s,t) = 0$ iff $(s,t) \notin G$.

It is trivial to see that: $\quad M^{\zeta_G} = M^G$

which means:

$$\forall f \in \mathbf{R}^S\ \forall t \in T: \sum_{s \in S} f(s).\zeta_G(s,t) = \sum_{s \in G^{-1}(t)} f(s)$$

Remark: For every finite set S, the set of weighted graphs $S \to S$, equipped with the operation of addition, with the real scalar multiplication and with the product is a **R**-algebra. In this **R**-algebra, the inverse (if it exists) of $\zeta_G$ is called the *Möbius function of G*, which is not to be confused with the Möbius transform of G.

**Lemma 1.** The sequence G

$$S = S_0 \xrightarrow{G_1} S_1 \xrightarrow{G_2} \ldots \xrightarrow{G_n} S_n = T$$

computes the Möbius transform of C(G) iff the sequence of weighted graphs

$$S = S_0 \xrightarrow{\zeta_{G_1}} S_1 \xrightarrow{\zeta_{G_2}} \ldots \xrightarrow{\zeta_{G_n}} S_n = T$$

computes the Möbius transform of $\zeta_{C(G)}$.

(and thus: $\zeta_{C(G)} = \zeta_{G_1}*\zeta_{G_2}* ... *\zeta_{G_n}$)

The following section examines the meaning of the equality $\zeta_{C(G)} = \zeta_{G_1}*\zeta_{G_2}* ... *\zeta_{G_n}$

### 1.3. Graph decomposition

The need for decomposing graphs (relatively to *) will become clear when we will see that a decomposition of a graph may decrease its *computational complexity*. In fact, the basic idea to getting a 'fast Möbius transform' is to decompose the inclusion relation of $\wp\Omega$.

**Definition 5.** Let G be the following sequence of graphs

$$S = S_0 \xrightarrow{G_1} S_1 \xrightarrow{G_2} \ldots \xrightarrow{G_n} S_n = T$$

a *path u of G* is a n-tuple $(g_1,...,g_n) \in G_1 \times ... \times G_n$ such that $\forall i \in \{1,...,n-1\}$: target($g_i$) = source($g_{i+1}$). The *source* of u is the source of $g_1$, the *target* of u is the target of $g_n$. Note

347

that contrary to the ordered pairs, the source and target of a path do not determine a path. The set of paths with source in $X \subseteq S$ and with target in $Y \subseteq T$ is denoted by $P_G(X,Y)$. For the sake of brevity $P_G(\{s\},\{t\})$ is replaced by $P_G(s,t)$. The following map: $P_G(S,T) \to G_n \circ ... \circ G_1$: $(g_1,...,g_n) \to g_n \circ ... \circ g_1 = g_1...g_n = (\mathrm{source}(g_1), \mathrm{target}(g_n))$ defines the usual *composition* or *product* of arrows. The inverse of the preceding map gives for each arrow $(s,t)$ the set of its *factorizations* of the form $g_1...g_n$ where $g_i \in G_i$.

**Lemma 2.** If G is the following sequence of graphs
$$S = S_0 \xrightarrow{G_1} S_1 \xrightarrow{G_2} ... \xrightarrow{G_n} S_n = T$$
then $(\zeta_{G_1} * \zeta_{G_2} * ... * \zeta_{G_n})(s,t) = \#P_G(s,t)$

Saying that every arrow g of $C(G)$ has a unique factorization of the form $g_1...g_n$ where $g_i \in G_i$ is equivalent to saying that $P_G(S,T) \to C(G)$: $(g_1,...,g_n) \to g_1...g_n$ is a bijection.

So, finally we get the fundamental theorem:

**Theorem 3.** Let G be the following sequence of graphs
$$S = S_0 \xrightarrow{G_1} S_1 \xrightarrow{G_2} ... \xrightarrow{G_n} S_n = T$$
G computes the Möbius transform of a graph
iff G computes the Möbius transform of the graph $C(G): S \to T$
iff $P_G(S,T) \to C(G)$: $(g_1,...,g_n) \to g_1...g_n$ is a bijection.

## 2. The Möbius transform of $(\wp\Omega, \subseteq)$

### 2.1 The fast Möbius transforms

Let us first recall the notion of the Hasse graph of a partial order relation.

**Definition 6.** If $(P, \leq)$ is a partially ordered set, then the *reflexive Hasse graph* of $\leq$ is $H(\leq) = \{(a,b) \in P \times P \mid a \leq b$ and $\forall x \in P: a < x \Rightarrow b \leq x\}: P \to P$. The *non-reflexive Hasse graph* of $(P, \leq)$ is $H(<) = \{(a,b) \in P \times P \mid a < b$ and $\forall x \in P: a < x \Rightarrow b \leq x\}$. The transitive closure of $H(\leq): P \to P$ is $\leq: P \to P$, and furthermore it is part of the folklore that (with respect to the inclusion relation) $H(\leq)$ is the smallest subgraph of $\leq$ which meets that property. Thus, $H(\leq)$ is characterized by the two properties:
(1) $T(H(\leq)) = \leq$ and (2) $T(G) = \leq \Rightarrow H(\leq) \subseteq G$.

We recall that G is a M-algorithm of $M^G$. In this section we describe other M-algorithms of $M^G$, and we show in section 4 that all these M-algorithms are optimal for a specific complexity measure. The next theorem provides a family of M-algorithms for computing the Möbius transform of $G=\{(X,Y) \in \wp\Omega \times \wp\Omega \mid X \subseteq Y\}$. We call these M-algorithms *the Hasse M-algorithms of G* or *the fast Möbius transforms of G*.

**Theorem 4.** If $\Omega = \{a_1, a_2, ..., a_n\}$, then the following M-algorithm H of length n:
$$\wp\Omega \xrightarrow{H_1} \wp\Omega \xrightarrow{H_2} ... \xrightarrow{H_n} \wp\Omega$$
where $H_i = \{(X,Y) \in \wp\Omega \times \wp\Omega \mid Y=X \text{ or } Y=X \cup \{a_i\}\}$
computes the Möbius transform of
$G=\{(X,Y) \in \wp\Omega \times \wp\Omega \mid X \subseteq Y\}$.

Note the fundamental fact: $U(H) = H(\subseteq) = H(G)$. That is the reason for also calling them the *Hasse* M-algorithms of G.

If we take into account the condition $X \neq \emptyset$, the previous theorem simply becomes:

**Theorem 4'.** If $\Omega = \{a_1, a_2, ..., a_n\}$, then the following M-algorithm H of length n:
$$\wp\Omega \xrightarrow{H_1} \wp\Omega \xrightarrow{H_2} ... \xrightarrow{H_n} \wp\Omega$$
where
$H_i = \{(X,Y) \in \wp\Omega \times \wp\Omega \mid X \neq \emptyset$ and $(Y=X$ or $Y=X \cup \{a_i\})\}$
computes the Möbius transform of
$\{(X,Y) \in \wp\Omega \times \wp\Omega \mid X \neq \emptyset, X \subseteq Y\}$.

**Example.** If $\Omega = \{a,b,c\}$, we have 6=3! different Hasse M-algorithms on $(\wp\Omega, \subseteq)$. Each total order on the set $\Omega$ determines a Hasse M-algorithm. Here are two of them:
$$\wp\Omega \xrightarrow{H_a} \wp\Omega \xrightarrow{H_b} \wp\Omega \xrightarrow{H_c} \wp\Omega \text{ and}$$
$$\wp\Omega \xrightarrow{H_c} \wp\Omega \xrightarrow{H_b} \wp\Omega \xrightarrow{H_a} \wp\Omega$$
where $H_a = \{(X,Y) \in \wp\Omega \times \wp\Omega \mid X \neq \emptyset$ and $(Y=X$ or $Y=X \cup \{a\})\}$ and similarly for $H_b$ and $H_c$. The last M-algorithm may be represented 'vertically' by:

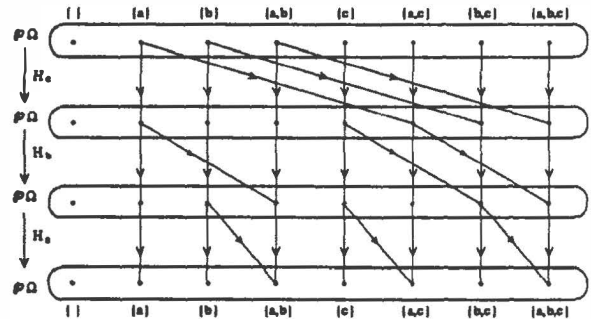

### 2.2 The inverse Möbius transform

It can easily be shown that the functional $M(G)$, determined by a graph $G: S \to T$, is not always injective so there is no hope, in the general case, to be able to 'reverse' (i.e. to get a left inverse of) M(G). That means that there does not always exist a functional F: T → S such that: F ∘ M(G) = 1 (the identity function on $R^S$). The problem of the existence of an inverse has a nice solution in the more general setting of weighted graphs. In fact, if $\zeta_G$ has an inverse (for the product *), then because M is a functor we get: $\zeta_G * (\zeta_G)^{-1} = \zeta_{1S}$ and so we have: $(M(\zeta_G))^{-1} \circ M(G) = 1$ (the identity function on $R^S$).

The Möbius inversion theorem gives a sufficient condition on the graph G for the functional $M(\zeta_G)$ to be a bijection.

**Möbius inversion theorem** [Aigner 79 p.141]. For every partial order relation graph G:S → S, there exists a weighted graph $\mu_G = (\zeta_G)^{-1}$ such that: $\zeta_G * \mu_G = \zeta_{1S} = \mu_G * \zeta_G$.

The weighted graph $\mu_G = (\zeta_G)^{-1}$ is called the *Möbius function of G*. $\mu_G: S \times S \to R : (s,t) \to \mu(s,t)$ is described recursively by:

$$\forall (s,t) \in S \times S: \mu_G(s,t) = -\sum_{s \le x < t} \mu_G(s,x) = -\sum_{s < x \le t} \mu_G(x,t)$$

with the following 'halting conditions':

$\forall s,t \in S: \mu_G(s,s) = 1$, and $\mu_G(s,t) = 0$ if $(s,t) \notin G$.

There also exists an iterative formula [Birkhoff 61 p.15] and [Aigner 79 p.146]:

$$\forall (s,t) \in S \times S: \mu_G(s,t) = \sum_{i=0}^{\infty} (-1)^i \lambda_i(s,t)$$

where $\lambda_i(s,t)$ = the number of chains of length i from s to t (i.e. totally ordered subsets of i+1 elements with s as minimum and t as maximum)

For our concern we simply need to find the inverse (with respect to *) of the graph $H_i = \{(X,Y) \in \wp\Omega \times \wp\Omega \mid Y=X$ or $Y = X \cup \{a_i\}\}$. In this case, it can readily be verified either directly or by using the Möbius inversion theorem that the following weighted graph is an inverse (hence *the* inverse) of $H_i$. $\mu_i: \wp\Omega \times \wp\Omega \to R$, defined by: $\forall X \in \wp\Omega: \mu_i(X,X) = 1, \mu_i(X, X \cup \{a_i\}) = -1$, else $\mu_i(X,Y) = 0$.

Thus we get:

**Theorem 5.** If $\Omega = \{a_1, a_2, ..., a_n\}$, then the following M-algorithm of weighted graphs computes the inverse Möbius transform of $(\wp\Omega, \subseteq)$:

$$\wp\Omega \xrightarrow{\mu_1} \wp\Omega \xrightarrow{\mu_2} ... \xrightarrow{\mu_n} \wp\Omega$$

where the $\mu_i: \wp\Omega \times \wp\Omega \to R$ are defined by:
$\forall X \in \wp\Omega: \mu_i(X,X) = 1, \mu_i(X, X \cup \{a_i\}) = -1$, else $\mu_i(X,Y) = 0$.

The M-algorithm can be represented vertically by: (The label on each arrow is its weight. For the simplification of the diagram we have not represented the identity arrows, which all have weight 1. By the way, all the other non-represented arrows have weight 0)

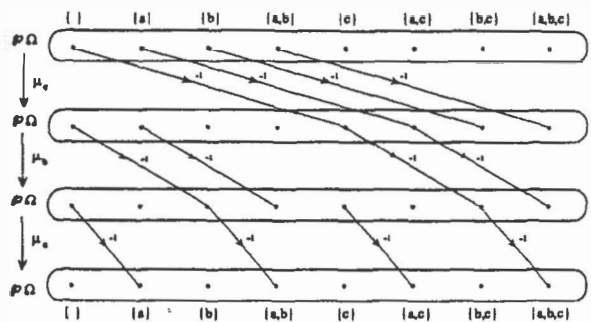

Since the number of arrows of weight (-1) in the path from X to A is #(A–X), the mass m(X) is multiplied by $(-1)^{\#(A-X)}$, and so we rediscover the classical formula transforming bel into m.

## 3. Computational complexity of sequences of graphs

We want to count up the number of additions performed by the algorithm on the 'worst possible input'. By 'worst possible input' we mean that 'a+b' stands for one addition whatever the values of a and b may be. No multiplication is needed in the case of M-algorithms of *graphs*. The 'cost' function will give the number of additions. We define it first for the vertices of a graph, then for a whole graph, and finally for M-algorithms of graphs.

**Definition 7.** Let G: S → T be a graph and t be an element of T.

(1) $\text{cost}_G(t) = \max\{0, \#G^{-1}(t) - 1\}$
    $= \#G^{-1}(t) - 1$ if $G^{-1}(t) \ne \emptyset$, else 0

(2) $\text{cost}(G) = \sum_{t \in T} \text{cost}_G(t)$
    $= \sum_{t \in T, G^{-1}(t) \ne \emptyset} (\#G^{-1}(t) - 1)$
    $= (\sum_{t \in T, G^{-1}(t) \ne \emptyset} \#G^{-1}(t)) - \#G(S) = \#G - \#G(S)$

(3) If A is a sequence of graphs
$$S = S_0 \xrightarrow{G_1} S_1 \xrightarrow{G_2} ... \xrightarrow{G_n} S_n = T$$
then $\text{cost}(A) = \sum_{i=1}^{n} \text{cost}(G_i)$

Examples:
1. Cost of the obvious M-algorithm G of
$\mathbf{M}\{(X,Y)\in \wp\Omega\times\wp\Omega \mid X\neq\emptyset, X\subseteq Y\}$

We get : $\text{cost}(G) = 3^n - 2^{n+1} + 1$, if $\#\Omega = n$

2. Cost of the Hasse M-algorithm H of
$\mathbf{M}\{(X,Y)\in \wp\Omega\times\wp\Omega \mid X\neq\emptyset, X\subseteq Y\}$

We get: $\text{cost}(H) = n\, 2^{n-1} - n$, if $\#\Omega = n$

Comparison between the two M-algorithms :

| $\#\Omega$ | $\#\wp\Omega$ | cost(G) | cost(H) | cost(G)/cost(H) |
|---|---|---|---|---|
| 5 | 32 | 180 | 75 | 2.4 |
| 8 | 256 | 6 050 | 1 016 | 5.9 |
| 10 | 1 024 | 57 002 | 5 110 | 11.1 |
| 12 | 4 096 | 523 250 | 24 564 | 21.3 |
| 15 | 32 768 | 14 283 372 | 245 745 | 58.1 |
| 20 | $10^6$ | $3.10^9$ | $10^7$ | 332.3 |

3. Cost of the Hasse M-algorithm of $M^{\subseteq}$
We get: $\text{cost}(H) = n\, 2^{n-1}$, if $\#\Omega = n$

## 4. Optimality of the fast Möbius transforms

The following theorem provides a lower bound for the complexity of the M-algorithms computing the Möbius transform of a finite lattice.

**Theorem 6.** Let $(L,\leq)$ be a finite lattice. If A is a M-algorithm of graphs computing $M^{\leq}$, then $\text{cost}(A) \geq \text{cost}(H(\leq))$.

In fact a slightly more general result can be proved:
**Theorem 7.** Let $(P,\leq)$ be a finite partially ordered set in which every upper bounded subset has a least upper bound. If A is a M-algorithm of graphs which computes $M^{\leq}$,
then $\text{cost}(A) \geq \text{cost}(H(\leq))$.

As a corollary of theorem 7 we get that the Hasse M-algorithms of $(\wp\Omega,\subseteq)$ are optimal among all M-algorithms computing $M^{\subseteq}$.

**Corollary 1.** The fast Möbius transforms of $(\wp\Omega,\subseteq)$ are optimal.

Let $\Omega = \{a_1,a_2,...,a_n\}$ and $(\wp\Omega,\subseteq)$ be the corresponding finite boolean lattice. If (1) A is a M-algorithm of graphs which computes $M^{\subseteq}$, and (2) H is a Hasse M-algorithm of $M^{\subseteq}$

$$\wp\Omega \xrightarrow{H_1} \wp\Omega \xrightarrow{H_2} \ldots \xrightarrow{H_n} \wp\Omega$$

where $H_i = \{(X,Y)\in \wp\Omega\times\wp\Omega \mid Y=X \text{ or } Y=X\cup\{a_i\}\}$,
then $\text{cost}(A) \geq \text{cost}(H)$.

We get the same result if we add the condition $X \neq \emptyset$.
Indeed:
**Corollary 2.** The fast Möbius transforms for D-S theory are optimal.

Let $\Omega = \{a_1,a_2,...,a_n\}$. If (1) A is a M-algorithm of graphs which computes the Möbius transform $M^G$ of $G=\{(X,Y)\in \wp\Omega\times\wp\Omega \mid X\neq\emptyset, X\subseteq Y\}$, and
(2) H is the following algorithm :

$$\wp\Omega \xrightarrow{H_1} \wp\Omega \xrightarrow{H_2} \ldots \xrightarrow{H_n} \wp\Omega$$

where $H_i = \{(X,Y)\in \wp\Omega\times\wp\Omega \mid X\neq\emptyset \text{ and } (Y=X \text{ or } Y=X\cup\{a_i\})\}$,
then $\text{cost}(A) \geq \text{cost}(H)$

Remark: the condition '*every upper bounded subset has a least upper bound*' cannot be removed otherwise we get counter-examples.

## 5. Application

### 5.1 Statement of the problem

An application of the previous techniques to the computation of Dempster rule of combination is shown in this last section. The framework is the Dempster-Shafer theory of evidence.
Most often, in DS theory, the easiest way to represent pieces of evidence is by using basic belief assignments, say $m_1$ and $m_2$. The mass distributions $m_1$ and $m_2$ are then combined together. Finally, the combined mass distribution $m_1 \otimes m_2$ is most easily interpreted when transformed back into its corresponding belief function and/or plausibility function. So, we intend to compute the transformation $(m_1,m_2) \to m_1\otimes m_2 \to \text{Pl}_{m_1\otimes m_2}$: first, by using the usual algorithms, and second by using the fast algorithms developed in the preceding sections. Eventually we compare the cost, both in additions and in multiplications, of the two computing ways. Let us first briefly recall the definitions of commonality function, plausibility function and Dempster's rule of combination.

### 5.2 Commonality functions

We have seen that the Möbius transform induced by the relation $\{(X,Y)\in \wp\Omega\times\wp\Omega \mid X\neq\emptyset, X\subseteq Y\}$ on $\wp\Omega$ is a functional which, in the D-S theory of evidence, is said to transform a *basic belief assignment* on $\wp\Omega$ into a *belief function*. There are other interesting Möbius transforms (induced by other relations!), one of which transforms a *basic belief assignment* into the so-called *commonality function*. Thus any basic belief assignment m determines its *commonality function* $Q_m$: $\wp\Omega \to [0\ 1]$ defined by:

$$\forall A \in \wp\Omega : Q_m(A) = \sum_{X \supseteq A} m(X)$$

$Q_m(A)$ is the *commonality number* of A induced by the bba m. In our general setting this transform is the Möbius transform induced by the relation $\{(X,Y) \in \wp\Omega \times \wp\Omega \mid X \supseteq Y\}$. All that has been said for the relations
$\{(X,Y) \in \wp\Omega \times \wp\Omega \mid X \neq \emptyset, X \subseteq Y\}$ and
$(X,Y) \in \wp\Omega \times \wp\Omega \mid X \subseteq Y\}$ applies as well for the relation defining the commonality function. So we get (by reversing the arrows) the

**Theorem 8.** If $\Omega = \{a_1, a_2, ..., a_n\}$, then the following M-algorithm H of length n :

$$\wp\Omega \xrightarrow{H_1} \wp\Omega \xrightarrow{H_2} \ldots \xrightarrow{H_n} \wp\Omega$$

where $H_i = \{(X,Y) \in \wp\Omega \times \wp\Omega \mid Y=X \text{ or } Y=X-\{a_i\}\}$ computes the Möbius transform of
$H = \{(X,Y) \in \wp\Omega \times \wp\Omega \mid X \supseteq Y\}$.
Moreover, any other M-algorithm G computing $M^H$ satisfies $cost(G) \geq cost(H)$.

The M-algorithm transforming a bba into its commonality function is: (the identity arrows, which all have weight 1, have not been represented)

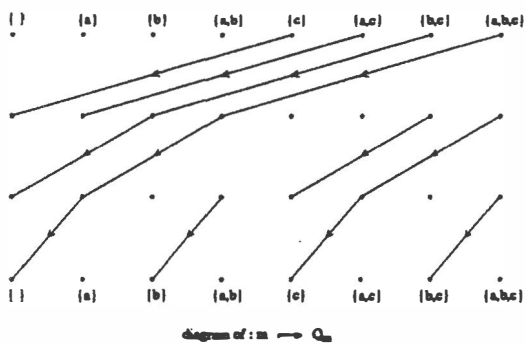

diagram of : m → $Q_m$

Let us, as an example, provide a 'real life' algorithm for this transform.

Given a total order on $\Omega = \{a_1, a_2, ..., a_n\}$, any subset of $\Omega$ is bijectively represented by a n-digit binary numeral. The bijection is realized in the following way: the element $a_i$ belongs to the subset A of $\Omega$ iff the $i^{th}$ digit of the numeral representing A is 'on' (i.e. is 1). Using this coding of the subsets of $\Omega$, it is not difficult to get the following algorithm:

**Input:** array $v[0:2^n-1]$   'contains the masses m'
**Output:** array $v[0:2^n-1]$   'contains the commonality function induced by m'

**Procedure** mtoq(v):   'transforms mass into commonality'
**begin**
  for i ← 1 step 1 until n do
    for j ← 1 step 2 until $2^i$ do
      for k ← 0 step 1 until $2^{n-i}-1$ do
        $v[(j-1).2^{n-i}+k] \leftarrow v[(j-1).2^{n-i}+k] + v[j.2^{n-i}+k]$
      end
    end
  end

### 5.3 Dempster's rule of combination

The significance of the commonality function lies the fact that it supplies a much simpler rule for computing Dempster's rule of combination. Let us first recall Dempster's rule of combination (we won't consider the normalizing factor here). This rule provides a bba $m_1 \otimes m_2$ given two bba, $m_1$ and $m_2$:

$$\forall A \in \wp\Omega : m_1 \otimes m_2(A) = \sum_{X \cap Y = A} m_1(X) \cdot m_2(Y)$$

The usefulness of the function Q is due to the following simple formula [Shafer 76 p.61]:

$$\forall A \in \wp\Omega : Q_{m_1 \otimes m_2}(A) = Q_{m_1}(A) \cdot Q_{m_2}(A)$$

### 5.4 Plausibility functions

Any basic belief assignment m determines its *plausibilty function* $Pl_m$: $\wp\Omega \to [0\ 1]$ defined by:

$$\forall A \in \wp\Omega : Pl_m(A) = \sum_{X \cap A \neq \emptyset} m(X)$$

It is proved in [Shafer 76 p.44] that:

$$Pl_Q(A) = \sum_{X \subseteq A, X \neq \emptyset} (-1)^{\#X+1} \cdot Q(X)$$

This last expression can be transformed into:

$$= abs\left( \sum_{X \subseteq A, X \neq \emptyset} (-1)^{\#(A-X)} \cdot Q(X) \right)$$

So, the transformation from the commonality function into the plausibility function is the same as the transformation from the belief function into the basic belief assignment – for which we can apply the fast algorithm – followed by the absolute value function.

An algorithm transforming a commonality function into its plausibility function is represented below (with the now usual omission of the identity arrows, which have weight 1). The last row of arrows represents the absolute value function.

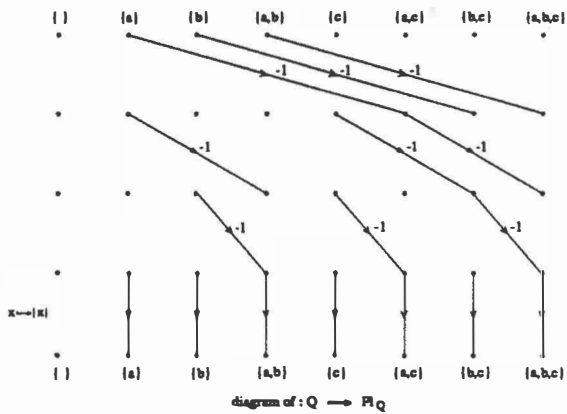

diagram of : $Q \to Pl_Q$

## 5.5 Comparison between the two computing ways

The problem at hand is to compute the transformation:
$(m_1, m_2) \to m_1 \otimes m_2 \to Pl_{m_1 \otimes m_2}$

The slow way and the fast way for computing the transformation may be represented together in the same following commutative diagram:

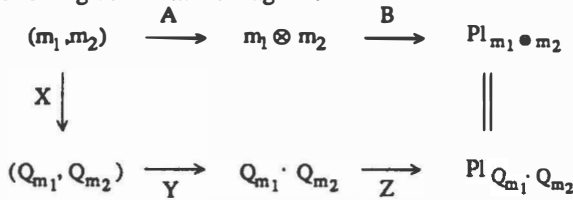

where the slow 'road' is AB and the fast 'road' is XYZ. As a matter of fact let us count the number of additions and the number of multiplications performed by each algorithm. ($n = \#\Omega$)

| Number of | additions | multiplications |
|---|---|---|
| obvious algorithm for (A): | $2^n(2^n-1)$ | $2^{2n}$ |
| obvious algorithm for (B): | $3^n-2^n$ | 0 |
| fast algorithm for (X): | $n \cdot 2^n$ | 0 |
| fast algorithm for (Y): | 0 | $2^n$ |
| fast algorithm for (Z): | $n(2^{n-1}-1)$ | 0 |

So, the two ratios *slow algorithm/fast algorithm* are respectively, for the number of additions and the number of multiplications, the following:

| n | $(3^n+2^{2n}-2^{n+1})/(n2^n+n2^{n-1}-n)$ | $2^n$ |
|---|---|---|
| 5 | 5 | 32 |
| 8 | 23 | 256 |
| 10 | 72 | 1 024 |
| 12 | 234 | 4 096 |
| 15 | 1 475 | 32 768 |
| 20 | 35 000 | $10^6$ |

The content of this section is a good illustration of what could be a taoist principle:

*If you are in a hurry, make a detour!*

## Conclusions

The generalized point of view adopted in this paper has allowed us to discover the fastest algorithms among a large class of algorithms computing the Möbius transform of a boolean lattice. As an application of these fast algorithms, we have shown how it can be used to compute other transforms of interest for the Dempster-Shafer theory.

All what has been stated in this paper can in fact be translated into the language of matrices, but such a translation would lose the conceptual insight provided by the 'graphic' framework. The same phenomenon is known to appear in linear algebra, where the framework of linear mappings between vector spaces indisputably provides greater insight into the matrix calculus.

The case of 'almost null' distributions which most frequently occurs in 'practical' uses of the Dempster-Shafer theory of evidence has not been discussed in this paper.